\documentclass[letterpaper, 10 pt, conference]{ieeeconf}  

\IEEEoverridecommandlockouts                              

\usepackage{graphics} 
\usepackage{epsfig} 
\usepackage{mathtools}
\usepackage{times} 
\usepackage{amsmath} 
\usepackage{amssymb}  
\usepackage{bm}
\usepackage{multirow}
\usepackage{makecell}
\usepackage{bbding}

\usepackage[utf8]{inputenc} 
\usepackage[T1]{fontenc}    
\usepackage{hyperref}       
            
\usepackage{url}            
\usepackage{booktabs}       
\usepackage{amsfonts}       
\usepackage{nicefrac}       
\usepackage{microtype}      
\usepackage{xcolor}         
\usepackage{cleveref}
\usepackage{physics}
\usepackage{hyperref}
\usepackage{graphicx}
\usepackage{float}
\usepackage{multirow}
\usepackage{diagbox}

\usepackage{algorithm}
\usepackage{algpseudocode}
\usepackage{arydshln}

\usepackage{multirow}
\usepackage{listings}
\usepackage{graphicx}
\usepackage{siunitx} 
            
\definecolor{dkgreen}{rgb}{0,0.6,0}
\definecolor{gray}{rgb}{0.5,0.5,0.5}
\definecolor{mauve}{rgb}{0.58,0,0.82}
\title{\LARGE \bf
TransDiff: Diffusion-Based Method for Manipulating Transparent Objects Using a Single RGB-D Image
}

\author{Haoxiao Wang$^{1*}$, Kaichen Zhou$^{1*}$, Binrui Gu$^{1}$, Zhiyuan Feng$^{2}$, Weijie Wang$^{3}$, Peilin Sun$^{3}$,\\ Yicheng Xiao$^{4}$, Jianhua Zhang$^{5}$, Hao Dong$^{1\dag}$
\thanks{* Equal contribution}
\thanks{$^{\dag}$ Corresponding author}%
\thanks{$^{1}$ Haoxiao Wang, Kaichen Zhou, Binrui Gu and Hao Dong are with CFCS, School of CS,
Peking University and National Key Laboratory for Multimedia Information Processing. 
$^{2}$ Zhiyuan Feng is with Tsinghua University. 
$^{3}$ Weijie Wang and Peilin Sun are with Zhejiang University. 
$^{4}$ Yicheng Xiao is with Southeast University. 
$^{5}$ Jianhua Zhang is with Tianjin University of Technology.}%
}

\begin{document}

\maketitle
\thispagestyle{empty}
\pagestyle{empty}

\begin{abstract}

Manipulating transparent objects presents significant challenges due to the complexities introduced by their reflection and refraction properties, which considerably hinder the accurate estimation of their 3D shapes.
To address these challenges, we propose a single-view RGB-D-based depth completion framework, TransDiff, that leverages the Denoising Diffusion Probabilistic Models(DDPM) to achieve material-agnostic object grasping in desktop.
Specifically, we leverage features extracted from RGB images, including semantic segmentation, edge maps, and normal maps, to condition the depth map generation process.
Our method learns an iterative denoising process that transforms a random depth distribution into a depth map, guided by initially refined depth information, ensuring more accurate depth estimation in scenarios involving transparent objects.
Additionally, we propose a novel training method to better align the noisy depth and RGB image features, which are used as conditions to refine depth estimation step by step. 
Finally, we utilized an improved inference process to accelerate the denoising procedure.
Through comprehensive experimental validation, we demonstrate that our method significantly outperforms the baselines in both synthetic and real-world benchmarks with acceptable inference time.  
The demo of our method can be found on: \url{https://wang-haoxiao.github.io/TransDiff/}

\end{abstract}

\section{INTRODUCTION}

Transparent objects, like glassware and plastic containers, are common in our daily lives and are used in various areas such as household tasks and laboratories. Although humans can easily see and interact with these objects, teaching robots to do the same is a big challenge. The main problem comes from the unique visual properties of transparent materials, like reflection and refraction, which make it hard for robotic systems to understand them. Traditional 3D sensors, such as LiDAR and RGB-D cameras, usually use depth data to create the shape of objects \cite{zhou2023manydepth2}. However, these sensors struggle to accurately capture the depth of transparent objects, often giving noisy or incomplete data \cite{jiang2023robotic}. This makes it difficult for robots to handle such objects effectively.

Recent studies have explored depth restoration and RGB-D-based grasping methods for handling transparent and specular objects. 
When it comes to reconstructing transparent objects, approaches can be categorized based on the number of viewpoints used to capture the object: (1) multi-view approaches; (2) single-view approaches. Works, e.g., Dex-NeRF \cite{ichnowski2021dex}, Evo-NeRF \cite{kerr2022evo}, GraspNeRF \cite{dai2023graspnerf}, leverage the generalizable neural radiance field (NeRF) \cite{mildenhall2021nerf,wang2021ibrnet,zhou2023dynpoint,zhou2024neural}, learning to recover missing depths and correcting wrong depths as a separate step prior to performing grasping, enabling material-agnostic object grasping in clutter. 
RGBGrasp \cite{liu2024rgbgrasp}, depends on a limited set of RGB views to perceive the 3D surroundings containing transparent and specular objects and achieve accurate grasping.
These multi-view methods are computationally expensive, require multiple views, and are sensitive to calibration errors. 
In particular, in certain challenging scenarios, such as objects placed against a wall, these methods may fail.
Some single-view approaches, e.g., ClearGrasp \cite{sajjan2020clear}, DepthGrasp \cite{tang2021depthgrasp}, A4T \cite{jiang2022a4t}, RFTrans \cite{tang2024rftrans}, uses deep convolutional networks to infer surface normals, masks of transparent surfaces, and occlusion boundaries.
Then reconstructs the 3D surfaces of transparent objects (missing depth region) via the global optimization algorithm \cite{zhang2018deep}.
However, these methods rely heavily on the accuracy of initial depth estimates or other geometric information, usually tailored to specific physical models, making them less flexible when dealing with different scenarios.

\begin{figure}[t]
\begin{center}
    \includegraphics[width=\linewidth]{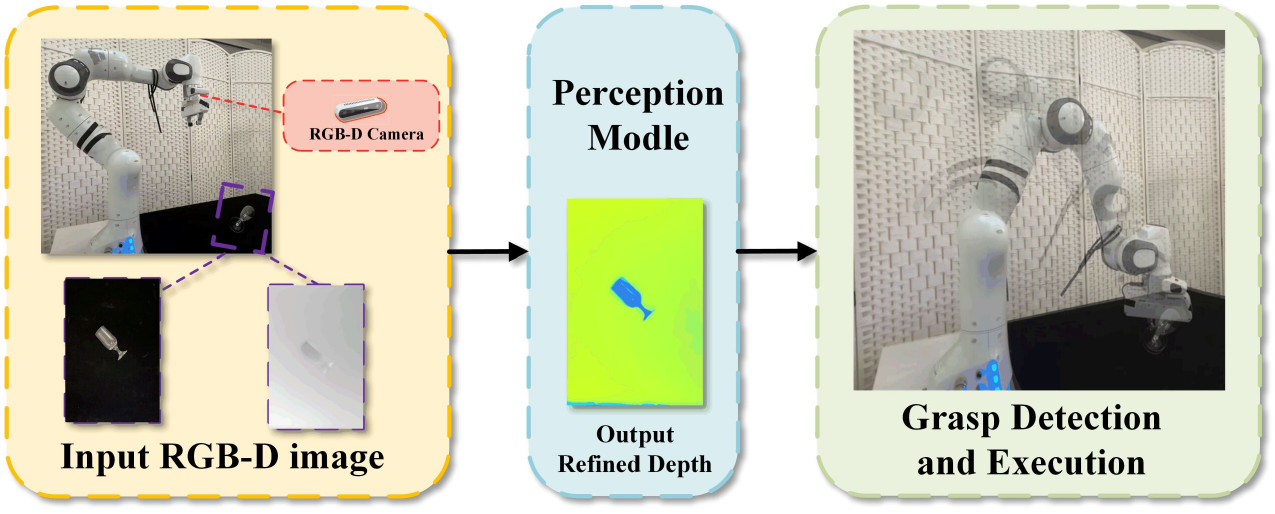}
\end{center}
\caption{\textbf{Overview of TransDiff.} We introduce a novel approach to reconstruct depth map of transparent object with single RGB-D image, learns an iterative denoising process that transforms a random depth distribution into a depth map. Once the point cloud image is obtained, we carry out grasp pose generation and grasp execution in the scene.}
\label{fig:teasor}
\vspace{-0.6cm}
\end{figure}

To mitigate these issues, we propose to leverage the diffusion models to solve the depth estimation for transparent objects.
This involves reformulating it as an iterative denoising process that generates the depth map from a random depth distribution.
The iterative refinement process allows the framework to capture both coarse and fine details of the scene at various stages.
In addition, diffusion models provide valuable benefits over networks trained with regression.
In particular, diffusion allows for approximate inference with multi-modal distributions, capturing uncertainty and ambiguity for transparent or specular objects.

This paper introduces TransDiff, a new depth completion framework for grasping transparent objects using a single-view RGB-D-based diffusion model.
Our approach takes in a random depth distribution as input and iteratively refines it through denoising steps guided by the initially refined depth map as visual conditions. 
The visual conditions are created by fusing feature maps from both the RGB image and the sparse depth map, combining texture and geometric information \cite{tang2024rftrans}.
This fusion enhances depth recovery for transparent objects by stabilizing the diffusion process and providing accurate guidance, even in challenging areas affected by transparency.
The denoising process is guided by visual conditions, merging it with the denoising block through a hierarchical structure.

The proposed TransDiff framework is evaluated on widely used public benchmarks ClearGrasp\cite{sajjan2020clear} and TranCG \cite{fang2022transcg}.
It could reach \textbf{0.032} and \textbf{0.028} RMSE on test split respectively on ClearGrasp and TranCG datasets, which exceeds state-of-the-art (SOTA) performance.
We conduct a detailed ablation study to gain a better understanding of the effectiveness and properties of the diffusion-based approach for completing the depth of transparent objects.
This study will explore the impact of various components and design choices on implementing the diffusion approach for transparent object depth completion.
In summary, our proposed diffusion-based method for manipulating transparent objects has many remarkable advantages over the previous state-of-the-art works:
\begin{itemize}
    \item We have developed a novel framework called TransDiff for completing depth in single RGB-D images. This framework addresses the diffusion-denoising problem iteratively using initially refined depth information as guidance. It enables the capture of uncertainty and ambiguity for transparent or specular objects.
    \item We design a feature fusion network that combines concatenation fusion and attention-based techniques to effectively utilize feature information from both RGB images and initially refined depth maps.
    \item Experimental results suggest TransDiff achieves state-of-the-art performance on both public benchmarks ClearGrasp and TranCG with acceptable inference costs.
\end{itemize}

\section{Related Work}

\subsection{Depth Estimation for Transparent Objects}
Estimating the 3D geometry of transparent objects poses a significant challenge due to the reflective and refractive nature of such materials.
Traditional approaches typically rely on structured light sensing\cite{curless1995better}, IR stereo\cite{alhwarin2014ir}, or cross-modal stereo techniques\cite{chiu2011improving} for transparent object detection. 
However, these methods are often constrained by specific setups and are challenging to integrate into real-world robotic applications. Recent advancements in deep learning models, a direct way is to restore depths before detecting grasp. ClearGrasp\cite{sajjan2020clear} first estimates geometric information from a single RGB image, then refines the depth through global optimization, enabling 3DoF transparent object grasping from a top-down perspective. 
A4T\cite{jiang2022a4t} introduces a hierarchical AffordanceNet, which is used for the first time to detect transparent objects and their associated affordances, encoding the relative positions of different parts of the object.
DepthGrasp\cite{tang2021depthgrasp} uses a generative adversarial network. The generator completes the depth maps by predicting the missing or inaccurate depth values, guided by the discriminator against the ground truth.
Other studies have shown the effectiveness of NeRFs as a promising way to represent scenes in several domains within depth reconstruction for transparent object grasping.
Neural radiance field (NeRF)\cite{mildenhall2021nerf} is an advanced neural network-based approach for scene representation and synthesis. Specifically, NeRF has shown potential in robot navigation\cite{sucar2021imap, adamkiewicz2022vision}, manipulation\cite{li20223d, dai2023graspnerf, ichnowski2021dex} and object
detection\cite{hu2023nerf}.
DexNeRF\cite{ichnowski2021dex} constructs a NeRF representation from 49 RGB images captured from multiple viewpoints, using volumetric rendering to produce a depth map from a top-down view for grasping tasks. GraspNeRF\cite{dai2023graspnerf}, on the other hand, improves NeRF's generalizability by reducing the input views to 6 sparse images, allowing the model to work on novel scenes without retraining.
However, these methods still rely on capturing images from a full 360-degree view around the scene to gather global context. This approach faces significant limitations in scenarios where the visual field is constrained, such as when grasping objects from a shelf, or when views from behind the scene are inaccessible. Additionally, these multi-view techniques come with high computational costs, are sensitive to calibration errors, and require numerous input images to function effectively.
In contrast, our TransDiff model addresses these shortcomings by requiring only a single-view RGB-D input for depth completion. This significantly reduces computational overhead and eliminates the need for capturing images from multiple angles. Furthermore, TransDiff leverages a diffusion model to refine depth estimation progressively, achieving robust performance even in challenging conditions where NeRF-based methods struggle. By aligning noisy depth information with RGB image features, our approach improves accuracy while maintaining efficiency, making it more practical for real-time robotic manipulation of transparent objects.

\begin{figure*}[t]
\centering
\includegraphics[width=1.0\textwidth]{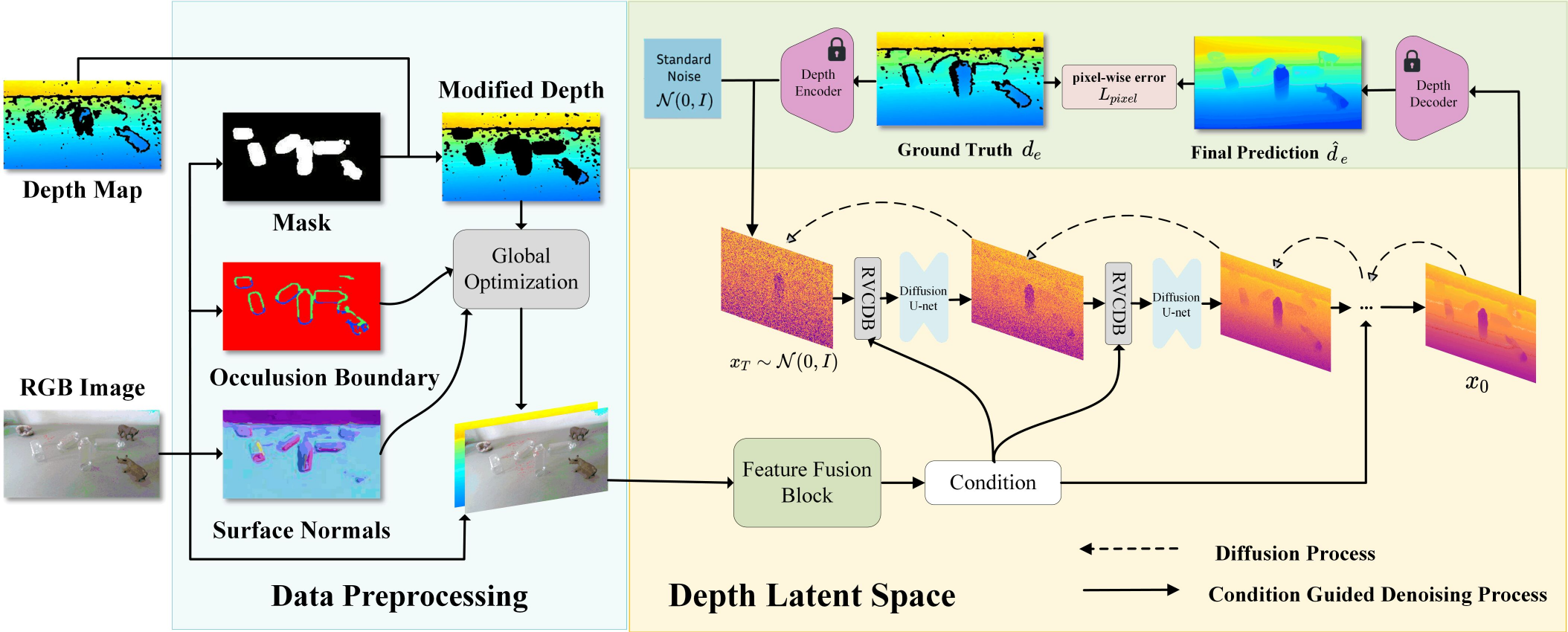} 
\caption{\textbf{TransDiff pipeline}
Given RGB-D image, TransDiff first predicts the mask, the boundary and the surface normal of transparent objects. Global optimization will then generate a depth map of the initial refinement, combined with an RGB image, and conduct feature fusion to integrate the features from both the RGB image and the initially refined depth map to create a combined feature representation. The Refined Visual Conditioned Denoising 
Block (RVCDB) iteratively refines the depth map through a denoising process guided by the fused visual conditions. 
The denoising process incorporates both the diffusion process and the guidance provided by the refined visual conditions, progressively improving the depth map's quality.
}
\label{fig2}
\end{figure*}

\subsection{Diffusion Models for Depth Estimation}
Diffusion models have gained significant attention for their generative capabilities, particularly in image generation tasks \cite{ho2020denoising, song2020score, dhariwal2021diffusion}. Initially, diffusion models were mainly explored for image synthesis, where they effectively learned to reverse a noise-injection process, gradually refining noisy images to generate high-quality outputs. More recently, their application in discriminative tasks such as object detection \cite{chen2023diffusiondet}, and image segmentation \cite{baranchuk2021label, graikos2022diffusion, wolleb2022diffusion} has emerged. DiffusionDet \cite{chen2023diffusiondet}, for example, extended the diffusion process to generate bounding box proposals, showcasing diffusion models' potential beyond image generation.
In depth estimation tasks, diffusion models have also demonstrated strong performance. Denoising Diffusion Probabilistic Models (DDPMs) \cite{ho2020denoising} are a powerful class of models that reverse a Gaussian noise diffusion process to recover data distributions, often leveraging latent spaces to reduce computational costs \cite{song2020denoising}. Conditional diffusion models extend DDPMs by conditioning on additional input data, such as images or semantic maps \cite{mirza2014conditional}. These models have been applied in depth estimation tasks, such as DDP \cite{ji2023ddp}, which encodes RGB images and decodes depth maps to achieve state-of-the-art results on datasets like KITTI. Other notable works include DiffusionDepth \cite{duan2023diffusiondepth}, which applies diffusion in the latent space, and DepthGen \cite{saxena2023monocular}, which handles noisy ground truth by employing a multi-task diffusion model.
Our method, TransDiff, leverages semantic segmentation, edge maps, and normal maps to get initially refined depth map as conditioning inputs, we progressively denoise and refine the depth maps, achieving robust performance across both synthetic and real-world scenarios. 

\section{Method}
In this section, we illustrate the details of our TransDiff framework for depth estimation of transparent objects.
We will first present the problem statement and then explore the architecture and key components of our proposed method.

\subsection{Problem Statement and Method Overview}

Given a single RGB-D image of transparent objects, Transdiff first performs a data preprocessing phase to get a initailly refined depth map and an rgb image. Specifically, a semantic segmentation will be obtained from the rgb image first, and this semantic segmentation will be processed in combination with the depth map to obtain modified depth. We continue to obtain edge maps and surface normals from rgb image through deep convolutional networks, among which surface normals are obtained through RFNet\cite{tang2024rftrans}, which uses refractive flow as an intermediate representation and avoids the limitations of directly predicting geometry from images, helping to bridge the sim2real gap.
Subsequently, we will fuse initailly refined depth map and rgb image information through feature fusion block to create a combined feature representation.
The Refined Visual Conditioned Denoising Block
(RVCDB) iteratively refines the depth map through a denoising process guided by the fused visual conditions. 
Through decoder, we can get the refined depth map of the transparent object from depth latent space.

The robot employs an eye-on-hand
configuration by strategically mounting one RGB-D camera on the wrist of the robot gripper. 
Finally, after we get the refined depth map, the corresponding point cloud map will be generated, then employ state-
of-the-art grasp pose detection methods, like GraspNet \cite{fang2020graspnet},
to leverage the reconstructed depth data for predicting optimal
grasp poses.

\subsection{Generative Formulation}

\subsubsection{DDPM Preliminaries}
Diffusion models are a class of latent variable models commonly used for generative tasks. The key idea is to iteratively degrade images by adding noise and train a neural network to reverse this process, effectively denoising the noisy images and generating high-quality outputs. The diffusion process typically follows a Markov chain, starting from clean data $\boldsymbol{x_0}$, and at each time step \( t \), Gaussian noise is progressively added to the data until the model reaches a fully noisy sample $\boldsymbol{x_T}$. The reverse process is then modeled by a neural network $\boldsymbol{\mu_\theta(x_t, t)}$, which predicts the denoised image step by step, eventually recovering the original image.

The forward diffusion process is described by the conditional probability distribution \( q(\boldsymbol{x_t}|\boldsymbol{x_0}) \), where \( \alpha_t \) represents the noise attenuation factor, and is defined as:

\begin{equation}
    q(\boldsymbol{x_t}|\boldsymbol{x_0}) := \mathcal{N}(\boldsymbol{x_t}|\sqrt{\bar{\alpha}_t}\boldsymbol{x_0}, (1 - \bar{\alpha}_t)\boldsymbol{I}),
\end{equation}

where \( \bar{\alpha}_t := \prod_{s=0}^t \alpha_s = \prod_{s=0}^t (1 - \beta_s) \), and \( \beta_s \) is the variance of the noise schedule. Over multiple iterations, the noisy sample $\boldsymbol{x_t}$ becomes less correlated with the original data $\boldsymbol{x_0}$, eventually reaching a highly noisy image $\boldsymbol{x_T}$.

The reverse diffusion process is handled by the neural network \( \boldsymbol{\mu_\theta}(\boldsymbol{x_t}, t) \), which predicts the denoised sample \( \boldsymbol{x_{t-1}} \) as:

\begin{equation}
    p_\theta(\boldsymbol{x_{t-1}}|\boldsymbol{x_t}) := \mathcal{N}(\boldsymbol{x_{t-1}}; \mu_\theta(\boldsymbol{x_t}, t), \sigma_t^2 \boldsymbol{I}),
\end{equation}

where \( \sigma_t^2 \) represents the variance of the transition. The goal is for the neural network to accurately predict $\boldsymbol{x_{t-1}}$ from $\boldsymbol{x_t}$, eventually reconstructing $\boldsymbol{x_0}$ from $\boldsymbol{x_T}$. During inference, the model begins with a random noise sample $\boldsymbol{x_T}$ and gradually denoises it to obtain the final image sample $\boldsymbol{x_0}$.


\subsubsection{Denoising as Depth Refinement}
Given input RGB-D image $\boldsymbol{d}$, 
the depth estimation task for transparent objects is normally formulated as $p(\boldsymbol{x} | \boldsymbol{d})$, where $\boldsymbol{x}$ is the targeted depth map of transparent objects. We reformulate the depth estimation as a refined-visual-condition guided denoising process which refines the depth distribution $\boldsymbol{x}_t$ iteratively as defined in Eq. 3 into the final depth map $x_0$.

\begin{equation}
p_\theta(\boldsymbol{x_{t-1}} | \boldsymbol{x_t}, \boldsymbol{d}) := \mathcal{N}(\boldsymbol{x_{t-1}}; \mu_\theta(\boldsymbol{x_t}, t, \boldsymbol{d}), \sigma_t^2 \boldsymbol{I}),    
\end{equation}

where model $\mu_\theta(\boldsymbol{x_t}, t, \boldsymbol{d})$ is trained to refine the depth latent $\boldsymbol{x_t}$ to $\boldsymbol{x_{t-1}}$. To speed up the denoising process, we used the enhanced inference process from DDIM \cite{song2020improved}, where $\sigma_t^2 I$ is set to 0 to make the prediction output deterministic.

\subsection{Network Architecture}
As shown in Fig.~\ref{fig2}, the network structure of TransDiff mainly consists of three parts. The first part is the data processing, through which the input RGB-D image is processed to obtain the initially refined depth map and rgb image. The second part is Refined Visual Conditioned Denoising Block
(RVCDB), witch iteratively refines the depth map through a denoising process guided by the fused visual conditions. The third part is Depth encoder and decoder.We will provide a thorough explanation in the upcoming content.

\subsubsection{Data Processing}
In depth completion tasks, it is common to incorporate surface normals and occlusion boundaries alongside visual features to supply 3D geometric details about objects. Surface normals help capture variations in local lighting, while occlusion boundaries are essential for identifying depth discontinuities. Specifically for transparent objects, Sajjan et al. \cite{sajjan2020clear} proposed the use of masks to discard uncertain depth values. \cite{tang2024rftrans} uses refractive flow as an intermediate representation to help reconstruct accurate surface geometry, thereby improving the accuracy of surface normal estimation. Following these approaches, we combine these three types of geometric information to generate an initially refined depth map through global optimization\cite{zhang2018deep}, along with the visual image, as inputs for our proposed TransDiff model.

\subsubsection{Refined Visual Conditioned Denoising Block}
Our TransDiff iteratively refines depth latent $\boldsymbol{x_t}$ and improves the prediction accuracy, guided by the fused visual information. Specifically, it is achieved by neural network model $\boldsymbol{\mu_}{\theta}(\boldsymbol{x_t}, t, \boldsymbol{c})$ which takes fused visual condition $\boldsymbol{c}$ and current depth latent $\boldsymbol{x_t}$ and predict the distribution $\boldsymbol{x_{t-1}}$. The fused visual condition $\boldsymbol{c} \in \mathbb{R}^{\frac{H}{4} \times \frac{W}{4} \times c}$ is constructed through multi-scale feature aggregation from both the RGB image and the initially refined depth map. The multi-scale aggregation ensures that both global and local information are captured effectively, enabling the model to handle various scales and fine details in the scene. These features are then projected into a lower resolution space through a local projection layer, ensuring that the dimensions of the visual condition match the current depth latent $\boldsymbol{x_t}$, while preserving the local relationships between features.

To enhance the ability of the model to focus on the most critical areas of the depth map, particularly in regions involving transparent objects, a self-attention mechanism \cite{vaswani2017attention} is applied to the fused features. The self-attention layer captures the relationships between different parts of the feature map, allowing the model to assign more weight to areas of high relevance. This helps in improving the precision of the depth refinement, especially in complex regions where standard depth estimation methods may struggle.
Following the attention mechanism, the fused depth latent is processed by a normal BottleNeck CNN layer and channel-wise attention with the residual connection.
The denoising output $\boldsymbol{x_{t-1}} $ is calculated by applying DDIM \cite{song2020score} inference process.

\subsubsection{Depth encoder and decoder.}

The frozen VAE encoder takes the input image and corresponding depth map, replicating the single-channel depth map into three channels to simulate an RGB image. The encoder then encodes the image into a latent space \cite{rombach2022high}. The depth map can be reconstructed from the encoded latent code With a minimal error. At inference, the average of the three decoded channels is taken as the predicted depth map.

\begin{figure*}[ht]
\centering
\includegraphics[width=1.0\textwidth]{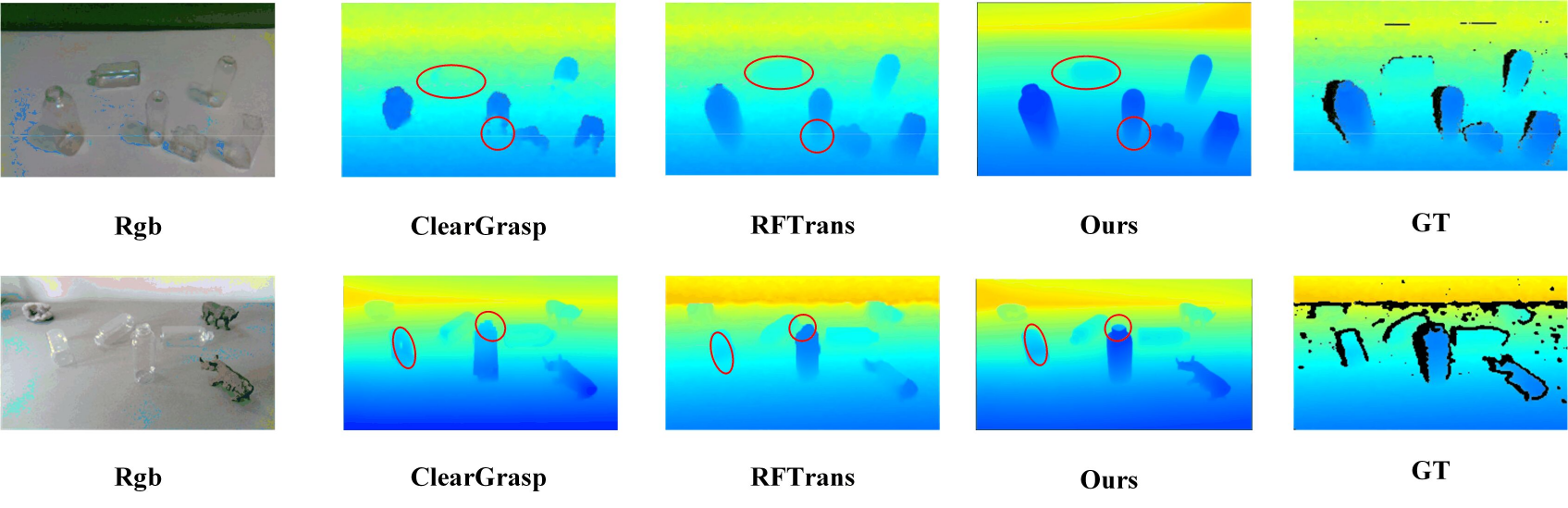} 
\caption{\textbf{Comparison Between TransDiff and Other methods in ClearGrasp Real-World Dataset.}
}
\label{fig3}
\end{figure*}

The depth latent $x_0 \in \mathbb{R}^{\frac{H}{2} \times \frac{W}{2} \times d}$ is passed through a decoder to obtain the final depth estimation. The depth decoder is composed of sequential $1 \times 1$ convolution, $3 \times 3$ de-convolution, and Sigmoid activation. 
The decoder and encoder are trained by minimizing the pixel-wise depth loss:

\begin{equation}
    L_{pixel} = \sqrt{\frac{1}{T} \sum_i \delta_i^2 + \frac{\lambda}{T^2} \left( \sum_i \delta_i \right)^2}
\end{equation}

where $\delta_i = \hat{d}_e - de$ is the pixel-wise depth error. The training process also incorporates L2 loss between the ground truth latent and the predicted latent depth.
The overall loss is a weighted sum of the individual losses:

\begin{equation}
L = \lambda_1 L_{ddim} + \lambda_2 L_{pixel} + \lambda_3 L_{2}    
\end{equation}

\subsection{Grasp Pose Detection and Manipulation}
In our TransDiff framework, after completing the depth refinement process, we utilize the resulting depth map  to facilitate grasp planning for transparent objects. Similar to prior works \cite{sajjan2020clear, jiang2022a4t, tang2024rftrans}, the refined point cloud serves as the input for a grasp planning algorithm. For this, we employ a state-of-the-art grasp model, such as GraspNet \cite{fang2020graspnet}, which directly predicts grasp poses from the depth data. The grasp proposals are evaluated based on an energy metric, with lower energy indicating a more stable and optimal grasp. This approach enables our system to seamlessly integrate with robotic manipulation tasks.

\section{Experiments}
In this section, we conduct a series of experiments to evaluate our diffusion method for transparent object depth reconstruction and manipulation. We benchmark our proposed TransDiff framework on both ClearGrasp and TransCG \cite{fang2022transcg} datasets, using the same evaluation protocols as in previous works.

\subsection{Depth Completion Experiments}
We compare our method with several representative approaches on both ClearGrasp and TransCG datasets. 

ClearGrasp is the first algorithm that utilizes deep learning and synthetic training data to estimate depth information for transparent objects. TranspareNet \cite{xu2021seeing} leverages existing depth information at the location of transparent objects to estimate the complete depth for transparent objects.
ImplicitDepth \cite{xu2021seeing} is the previous state-of-the-art depth completion method.
We also evaluate the concurrent work RFTrans \cite{tang2024rftrans} in experiments. All baselines are trained on ClearGrasp or TransCG datasets using their released source codes and optimal hyper-parameters for fair comparisons.

For depth estimation, we report standard metrics, including Root Mean Squared Error (RMSE), Absolute Relative Difference (REL), Mean Absolute Error (MAE), and the percentage of pixels with predicted depths satisfying various threshold values (\(\delta = 1.05, 1.10, 1.25\)) 
\begin{itemize}
    \item \textbf{RMSE}: the square root of the mean squared error between the predicted depths and the ground-truth depths.
    \item \textbf{REL}: the mean absolute relative error.
    \item \textbf{MAE}: the average absolute error between the predicted depths and the ground-truth values.
    \item \textbf{Threshold} $\delta$: the proportion of pixels for which the predicted depths meet the condition $\max(d/d^*, d^*/d) < \delta$, where $d$ and $d^*$ are the corresponding pixels in $\mathbf{D}$ and $\mathbf{D}^*$, respectively. The values for $\delta$ are set to 1.05, 1.10, and 1.25.
\end{itemize}

For all methods, the images are scaled to 320 $\times$ 240 during both training and testing. Training and testing are conducted using 8 NVIDIA GeForce RTX 3090 GPUs, and the average time to train an epoch is approximately 90 minutes.

\begin{table}[ht]
    \centering
    \caption{Quantitative comparison of methods on depth completion metrics on ClearGrasp dataset. Lower is better for RMSE, REL, and MAE. Higher is better for $\delta$ thresholds.}
    \resizebox{1\linewidth}{!}{
    \begin{tabular}{l|ccc|ccc}
    \hline
         & \multicolumn{3}{c}{Error Metrics} & \multicolumn{3}{c}{Accuracy Metrics} \\ 
    \hline
    \textbf{Method} & \textbf{RMSE$\downarrow$} & \textbf{REL$\downarrow$} & \textbf{MAE$\downarrow$} & \boldmath$\delta_{1.05} \uparrow$ & \boldmath$\delta_{1.10} \uparrow$ & \boldmath$\delta_{1.25} \uparrow$ \\
    \hline
    \textbf{TranDiff(Ours)} & \textbf{0.032} & \textbf{0.051} & \textbf{0.027} & \textbf{72.42} & \textbf{88.22} & 97.84
    \\
    RFTrans & 0.041 & 0.065 & 0.033 & 67.92 & 85.44 & \textbf{97.96}
    \\
    ClearGrasp & 0.048 & 0.076 & 0.039 & 53.23 & 79.01 & 95.94 \\
    ImplicitDepth  & 0.050 & 0.081 & 0.043 & 38.79 & 62.55      & 92.17
    \\
    \hline
    \end{tabular}
    }
\label{tab:cleargrasp}
\end{table}

\begin{table}[ht]
    \centering
    \caption{Quantitative comparison of methods on depth completion metrics on TransCG dataset. Lower is better for RMSE, REL, and MAE. Higher is better for $\delta$ thresholds.}
    \resizebox{1\linewidth}{!}{
    \begin{tabular}{l|ccc|ccc}
    \hline
         & \multicolumn{3}{c}{Error Metrics} & \multicolumn{3}{c}{Accuracy Metrics} \\ 
    \hline
    \textbf{Method} & \textbf{RMSE$\downarrow$} & \textbf{REL$\downarrow$} & \textbf{MAE$\downarrow$} & \boldmath$\delta_{1.05} \uparrow$ & \boldmath$\delta_{1.10} \uparrow$ & \boldmath$\delta_{1.25} \uparrow$ \\
    \hline
    \textbf{TranDiff(Ours)} & \textbf{0.028} & \textbf{0.042} & \textbf{0.019} & \textbf{81.45} & \textbf{94.81} & \textbf{99.72} 
    \\
    ClearGrasp & 0.052 & 0.081 & 0.038 & 51.48 & 74.02 & 97.19 \\
    TranspareNet  & 0.035 & 0.048 & 0.030 & 77.72 & 92.04      & 98.91
    \\
    \hline
    \end{tabular}
    }
\label{tab:transcg}
\end{table}

The quantitative results are reported in Tab. \ref{tab:cleargrasp} and Tab. \ref{tab:transcg}, and the qualitative results are visualized in Fig. \ref{fig3}.
In Tab. \ref{tab:cleargrasp}, results show that the proposed method can produce the lowest 
table mean error and it performs significantly better than others even on the most strict $\delta_{1.05}$ metric.
ImplicitDepth \cite{zhu2021rgb} utilize a local implicit neural representation based on ray-voxel pairs to achieve fast inference and iteratively improve depth estimation through a self-correcting refinement model.
ClearGrasp achieves competitive performance compared with the aforementioned approaches, benefitting from the global optimization functions and geometric information estimated from color images. In Tab. \ref{tab:transcg}, our method outperforms ClearGrasp and TranspareNet \cite{xu2021seeing} on both metrics and quality.

\subsection{Ablation Study}
In this section, we evaluate the effect of different inference settings, analyze the effects of conditioned denoising block.

\subsubsection{\textbf{Ablation Study on Denoising Inference}}
To further expose the properties of different inference steps, we conduct an ablation study on various inference settings.
We consider two strategies, one is to train with 1000 diffusion steps and the second is to train with different inference steps.
The ablation is conducted on the ClearGrasp dataset, where the variations of the metrics are reported in Tab. \ref{tab:ablation_study}.
We use 1000 diffusion steps during training, and directly altering inference steps significantly degrades performance, unlike in detection tasks where steps can be adjusted post-training. This is likely due to the detailed nature of depth maps, making the task more generative. However, by training with the target inference setting, we achieve faster inference with minimal performance loss.
\begin{table}[ht]
\centering
\caption{Ablation study on different inference settings on cleargrasp dataset, $t$ denotes the inference step.}
 \resizebox{1\linewidth}{!}{
\begin{tabular}{l|ccc|ccc}
\hline
\textbf{Method} & \textbf{Rel $\downarrow$} & \textbf{RMSE $\downarrow$} & \textbf{MAE $\downarrow$} & \boldmath$\delta_{1.05} \uparrow$ & \boldmath$\delta_{1.10} \uparrow$ & \boldmath$\delta_{1.25} \uparrow$ \\
\hline
\textbf{t=20} & \textbf{0.051} & \textbf{0.032} & \textbf{0.027} & \textbf{72.42} & 88.22 & \textbf{97.84} \\
\hline
\multicolumn{7}{c}{Directly change inference without training.} \\
\hline
t=15 & 0.086 & 0.045 & 0.064 & 69.86 & 86.73 & 96.92 \\
t=10 & 0.127 & 0.075 & 0.103 & 50.64 & 81.92 & 94.98 \\
t=5  & 0.192 & 0.117 & 0.154 & 38.04 & 70.85 & 88.25 \\
t=2  & 0.254 & 0.184 & 0.216 & 28.08 & 55.04 & 78.25 \\
\hline
\multicolumn{7}{c}{Train with different inference steps.} \\
\hline
t=15 & 0.079 & 0.042 & 0.054 & 71.26 & \textbf{88.89} & 97.29 \\
t=10 & 0.081 & 0.043 & 0.059 & 69.45 & 86.90 & 96.85 \\
t=5  & 0.083 & 0.045 & 0.061 & 67.87 & 85.12 & 96.56 \\
t=2  & 0.101 & 0.056 & 0.073 & 62.92 & 84.01 & 95.97 \\
\hline
\end{tabular}
}
\label{tab:ablation_study}
\end{table}

\subsubsection{\textbf{Ablation Study on  conditioned denoising block}}
For the design of conditioned denoising block, there are two main strategies: one is to directly extract visual condition from rgb image and add it to the denoising block; the other is leverage
features extracted from RGB images, including semantic
segmentation, edge maps, and normal maps, to condition the
depth map generation process. Intuitively, the latter seems more effective. The results are shown in 
Tab. \ref{tab:diffusion_methods}. It can be seen that the refined visual conditioned denoising block has a much better effect, even an order of magnitude advantage, because geometric information such as surface normal plays a crucial role in depth recovery of transparent objects.

\begin{table}[ht]
\centering
\caption{Ablation on different diffusion methods. Both methods are evaluated on offline official splits.}
\resizebox{1\linewidth}{!}{
\begin{tabular}{l|ccc|ccc}
\hline
\textbf{Method} & \textbf{Rel $\downarrow$} & \textbf{RMSE $\downarrow$} & \textbf{MAE $\downarrow$} & \boldmath$\delta_{1.05} \uparrow$ & \boldmath$\delta_{1.10} \uparrow$ & \boldmath$\delta_{1.25} \uparrow$ \\
\hline
\multicolumn{7}{c}{\textbf{ClearGrasp Dataset}} \\
\hline
\textbf{Refined} & 0.051 & 0.032 & 0.027 & 72.42 & 88.22 & 97.84 \\
only-rgb & 0.372 & 0.246 & 0.150 & 14.01 & 32.97 & 61.26 \\
\hline
\multicolumn{7}{c}{\textbf{TransCG Dataset}} \\
\hline
\textbf{Refined} & 0.042 & 0.028 & 0.019 & 81.45 & 94.81 & 99.72 \\
only-rgb & 0.384 & 0.224 & 0.179 & 16.44 & 36.25 & 68.97 \\
\hline
\end{tabular}
}
\label{tab:diffusion_methods}
\end{table}

\subsection{Simulation Grasping Experiments}
We train and evaluate our method
in Pybullet \cite{coumans2016pybullet}, which provides a versatile and accessible
platform for simulating the grasping task.
We measure our performance with the following evaluation metrics:
\begin{itemize}
    \item \textbf{Success Rate (SR)} The number of successful grasp divided by the number of total grasp attempts
    \item \textbf{Declutter Rate (DR)} The percentage of removed objects among all objects in the scene. The reported declutter rates are averages over all rounds. 
\end{itemize}
We collect 8 glass objects for grasping, and we set up 120 scenes for evaluation. In each scene, we randomly place 1 to 5 transparent objects. In each attempt, the robot should pick up an object according to the grasp pose generated by the grasp planning algorithm, GraspNet \cite{fang2020graspnet}. The results are shown in Tab. \ref{tab:success_rate}.

\begin{table}[ht]
\centering
\caption{Success rate of manipulation.}
\begin{tabular}{l|c|c}
\hline
\textbf{Method} & \textbf{SR (\%)} & \textbf{DR (\%)} \\
\hline
\textbf{TransDiff(ours)} & 87.5 & 86.70  \\
ClearGrasp    & 55.83 & 49.72  \\
RFTrans       & 75.83 & 69.24 \\
\hline
\end{tabular}
\label{tab:success_rate}
\end{table}

\subsection{Real-world Experiment}
We used a Franka Panda robot arm as our manipulator and attached a Realsense camera with RGB-only output to the robot’s end-effector. The camera observations are directly utilized by the agent without any processing. Videos can be found on \url{https://wang-haoxiao.github.io/TransDiff/}

\section{CONCLUSIONS}
In this work, we introduce TransDiff, a novel single-view RGB-D-based depth completion framework specifically designed for manipulating transparent objects. Our approach leverages Denoising Diffusion Probabilistic Models to refine depth estimation in complex scenarios involving reflection and refraction, enabling material-agnostic object grasping. By integrating some geometry features, we enhance the conditioning of the depth generation process and propose a new training method that better aligns noisy depth and RGB image features. Additionally, we improve the inference process to accelerate depth estimation. Extensive experiments on both synthetic and real-world datasets demonstrate the superiority of our method over existing baselines in terms of accuracy and efficiency.

\section{Acknowledgment}
This paper was supported by the National Youth Talent Support Program (8200800081) and National Natural Science Foundation of China (No. 62376006).

\newpage

\bibliographystyle{IEEEtran}
\bibliography{IEEEabrv,reference}

\begin{thebibliography}{10}
\providecommand{\url}[1]{#1}
\csname url@rmstyle\endcsname
\providecommand{\newblock}{\relax}
\providecommand{\bibinfo}[2]{#2}
\providecommand\BIBentrySTDinterwordspacing{\spaceskip=0pt\relax}
\providecommand\BIBentryALTinterwordstretchfactor{4}
\providecommand\BIBentryALTinterwordspacing{\spaceskip=\fontdimen2\font plus
\BIBentryALTinterwordstretchfactor\fontdimen3\font minus \fontdimen4\font\relax}
\providecommand\BIBforeignlanguage[2]{{%
\expandafter\ifx\csname l@#1\endcsname\relax
\typeout{** WARNING: IEEEtran.bst: No hyphenation pattern has been}%
\typeout{** loaded for the language `#1'. Using the pattern for}%
\typeout{** the default language instead.}%
\else
\language=\csname l@#1\endcsname
\fi
#2}}

\bibitem{zhou2023manydepth2}
K.~Zhou, J.~Bian, Q.~Xie, J.~Zheng, N.~Trigoni, and A.~Markham, ``Manydepth2: Motion-aware self-supervised monocular depth estimation in dynamic scenes,'' \emph{arXiv preprint arXiv:2312.15268}, 2023.

\bibitem{jiang2023robotic}
J.~Jiang, G.~Cao, J.~Deng, T.-T. Do, and S.~Luo, ``Robotic perception of transparent objects: A review,'' \emph{IEEE Transactions on Artificial Intelligence}, 2023.

\bibitem{ichnowski2021dex}
J.~Ichnowski, Y.~Avigal, J.~Kerr, and K.~Goldberg, ``Dex-nerf: Using a neural radiance field to grasp transparent objects,'' \emph{arXiv preprint arXiv:2110.14217}, 2021.

\bibitem{kerr2022evo}
J.~Kerr, L.~Fu, H.~Huang, Y.~Avigal, M.~Tancik, J.~Ichnowski, A.~Kanazawa, and K.~Goldberg, ``Evo-nerf: Evolving nerf for sequential robot grasping of transparent objects,'' in \emph{6th annual conference on robot learning}, 2022.

\bibitem{dai2023graspnerf}
Q.~Dai, Y.~Zhu, Y.~Geng, C.~Ruan, J.~Zhang, and H.~Wang, ``Graspnerf: Multiview-based 6-dof grasp detection for transparent and specular objects using generalizable nerf,'' in \emph{2023 IEEE International Conference on Robotics and Automation (ICRA)}.\hskip 1em plus 0.5em minus 0.4em\relax IEEE, 2023, pp. 1757--1763.

\bibitem{mildenhall2021nerf}
B.~Mildenhall, P.~P. Srinivasan, M.~Tancik, J.~T. Barron, R.~Ramamoorthi, and R.~Ng, ``Nerf: Representing scenes as neural radiance fields for view synthesis,'' \emph{Communications of the ACM}, vol.~65, no.~1, pp. 99--106, 2021.

\bibitem{wang2021ibrnet}
Q.~Wang, Z.~Wang, K.~Genova, P.~P. Srinivasan, H.~Zhou, J.~T. Barron, R.~Martin-Brualla, N.~Snavely, and T.~Funkhouser, ``Ibrnet: Learning multi-view image-based rendering,'' in \emph{Proceedings of the IEEE/CVF conference on computer vision and pattern recognition}, 2021, pp. 4690--4699.

\bibitem{zhou2023dynpoint}
K.~Zhou, J.-X. Zhong, S.~Shin, K.~Lu, Y.~Yang, A.~Markham, and N.~Trigoni, ``Dynpoint: Dynamic neural point for view synthesis,'' \emph{Advances in Neural Information Processing Systems}, vol.~36, pp. 69\,532--69\,545, 2023.

\bibitem{zhou2024neural}
K.~Zhou, ``Neural surface reconstruction from sparse views using epipolar geometry,'' \emph{arXiv preprint arXiv:2406.04301}, 2024.

\bibitem{liu2024rgbgrasp}
C.~Liu, K.~Shi, K.~Zhou, H.~Wang, J.~Zhang, and H.~Dong, ``Rgbgrasp: Image-based object grasping by capturing multiple views during robot arm movement with neural radiance fields,'' \emph{IEEE Robotics and Automation Letters}, 2024.

\bibitem{sajjan2020clear}
S.~Sajjan, M.~Moore, M.~Pan, G.~Nagaraja, J.~Lee, A.~Zeng, and S.~Song, ``Clear grasp: 3d shape estimation of transparent objects for manipulation,'' in \emph{2020 IEEE international conference on robotics and automation (ICRA)}.\hskip 1em plus 0.5em minus 0.4em\relax IEEE, 2020, pp. 3634--3642.

\bibitem{tang2021depthgrasp}
Y.~Tang, J.~Chen, Z.~Yang, Z.~Lin, Q.~Li, and W.~Liu, ``Depthgrasp: Depth completion of transparent objects using self-attentive adversarial network with spectral residual for grasping,'' in \emph{2021 IEEE/RSJ International Conference on Intelligent Robots and Systems (IROS)}.\hskip 1em plus 0.5em minus 0.4em\relax IEEE, 2021, pp. 5710--5716.

\bibitem{jiang2022a4t}
J.~Jiang, G.~Cao, T.-T. Do, and S.~Luo, ``A4t: Hierarchical affordance detection for transparent objects depth reconstruction and manipulation,'' \emph{IEEE Robotics and Automation Letters}, vol.~7, no.~4, pp. 9826--9833, 2022.

\bibitem{tang2024rftrans}
T.~Tang, J.~Liu, J.~Zhang, H.~Fu, W.~Xu, and C.~Lu, ``Rftrans: Leveraging refractive flow of transparent objects for surface normal estimation and manipulation,'' \emph{IEEE Robotics and Automation Letters}, 2024.

\bibitem{zhang2018deep}
Y.~Zhang and T.~Funkhouser, ``Deep depth completion of a single rgb-d image,'' in \emph{Proceedings of the IEEE conference on computer vision and pattern recognition}, 2018, pp. 175--185.

\bibitem{fang2022transcg}
H.~Fang, H.-S. Fang, S.~Xu, and C.~Lu, ``Transcg: A large-scale real-world dataset for transparent object depth completion and a grasping baseline,'' \emph{IEEE Robotics and Automation Letters}, vol.~7, no.~3, pp. 7383--7390, 2022.

\bibitem{curless1995better}
B.~Curless and M.~Levoy, ``Better optical triangulation through spacetime analysis,'' in \emph{Proceedings of IEEE International Conference on Computer Vision}.\hskip 1em plus 0.5em minus 0.4em\relax IEEE, 1995, pp. 987--994.

\bibitem{alhwarin2014ir}
F.~Alhwarin, A.~Ferrein, and I.~Scholl, ``Ir stereo kinect: improving depth images by combining structured light with ir stereo,'' in \emph{PRICAI 2014: Trends in Artificial Intelligence: 13th Pacific Rim International Conference on Artificial Intelligence, Gold Coast, QLD, Australia, December 1-5, 2014. Proceedings 13}.\hskip 1em plus 0.5em minus 0.4em\relax Springer, 2014, pp. 409--421.

\bibitem{chiu2011improving}
W.~W.-C. Chiu, U.~Blanke, and M.~Fritz, ``Improving the kinect by cross-modal stereo.'' in \emph{Bmvc}, vol.~1, no.~2.\hskip 1em plus 0.5em minus 0.4em\relax Dundee, 2011, p.~3.

\bibitem{sucar2021imap}
E.~Sucar, S.~Liu, J.~Ortiz, and A.~J. Davison, ``imap: Implicit mapping and positioning in real-time,'' in \emph{Proceedings of the IEEE/CVF International Conference on Computer Vision}, 2021, pp. 6229--6238.

\bibitem{adamkiewicz2022vision}
M.~Adamkiewicz, T.~Chen, A.~Caccavale, R.~Gardner, P.~Culbertson, J.~Bohg, and M.~Schwager, ``Vision-only robot navigation in a neural radiance world,'' \emph{IEEE Robotics and Automation Letters}, vol.~7, no.~2, pp. 4606--4613, 2022.

\bibitem{li20223d}
Y.~Li, S.~Li, V.~Sitzmann, P.~Agrawal, and A.~Torralba, ``3d neural scene representations for visuomotor control,'' in \emph{Conference on Robot Learning}.\hskip 1em plus 0.5em minus 0.4em\relax PMLR, 2022, pp. 112--123.

\bibitem{hu2023nerf}
B.~Hu, J.~Huang, Y.~Liu, Y.-W. Tai, and C.-K. Tang, ``Nerf-rpn: A general framework for object detection in nerfs,'' in \emph{Proceedings of the IEEE/CVF Conference on Computer Vision and Pattern Recognition}, 2023, pp. 23\,528--23\,538.

\bibitem{ho2020denoising}
J.~Ho, A.~Jain, and P.~Abbeel, ``Denoising diffusion probabilistic models,'' \emph{Advances in neural information processing systems}, vol.~33, pp. 6840--6851, 2020.

\bibitem{song2020score}
Y.~Song, J.~Sohl-Dickstein, D.~P. Kingma, A.~Kumar, S.~Ermon, and B.~Poole, ``Score-based generative modeling through stochastic differential equations,'' \emph{arXiv preprint arXiv:2011.13456}, 2020.

\bibitem{dhariwal2021diffusion}
P.~Dhariwal and A.~Nichol, ``Diffusion models beat gans on image synthesis,'' \emph{Advances in neural information processing systems}, vol.~34, pp. 8780--8794, 2021.

\bibitem{chen2023diffusiondet}
S.~Chen, P.~Sun, Y.~Song, and P.~Luo, ``Diffusiondet: Diffusion model for object detection,'' in \emph{Proceedings of the IEEE/CVF international conference on computer vision}, 2023, pp. 19\,830--19\,843.

\bibitem{baranchuk2021label}
D.~Baranchuk, I.~Rubachev, A.~Voynov, V.~Khrulkov, and A.~Babenko, ``Label-efficient semantic segmentation with diffusion models,'' \emph{arXiv preprint arXiv:2112.03126}, 2021.

\bibitem{graikos2022diffusion}
A.~Graikos, N.~Malkin, N.~Jojic, and D.~Samaras, ``Diffusion models as plug-and-play priors,'' \emph{Advances in Neural Information Processing Systems}, vol.~35, pp. 14\,715--14\,728, 2022.

\bibitem{wolleb2022diffusion}
J.~Wolleb, R.~Sandk{\"u}hler, F.~Bieder, P.~Valmaggia, and P.~C. Cattin, ``Diffusion models for implicit image segmentation ensembles,'' in \emph{International Conference on Medical Imaging with Deep Learning}.\hskip 1em plus 0.5em minus 0.4em\relax PMLR, 2022, pp. 1336--1348.

\bibitem{song2020denoising}
J.~Song, C.~Meng, and S.~Ermon, ``Denoising diffusion implicit models,'' \emph{arXiv preprint arXiv:2010.02502}, 2020.

\bibitem{mirza2014conditional}
M.~Mirza and S.~Osindero, ``Conditional generative adversarial nets,'' \emph{arXiv preprint arXiv:1411.1784}, 2014.

\bibitem{ji2023ddp}
Y.~Ji, Z.~Chen, E.~Xie, L.~Hong, X.~Liu, Z.~Liu, T.~Lu, Z.~Li, and P.~Luo, ``Ddp: Diffusion model for dense visual prediction,'' in \emph{Proceedings of the IEEE/CVF International Conference on Computer Vision}, 2023, pp. 21\,741--21\,752.

\bibitem{duan2023diffusiondepth}
Y.~Duan, X.~Guo, and Z.~Zhu, ``Diffusiondepth: Diffusion denoising approach for monocular depth estimation,'' \emph{arXiv preprint arXiv:2303.05021}, 2023.

\bibitem{saxena2023monocular}
S.~Saxena, A.~Kar, M.~Norouzi, and D.~J. Fleet, ``Monocular depth estimation using diffusion models,'' \emph{arXiv preprint arXiv:2302.14816}, 2023.

\bibitem{fang2020graspnet}
H.-S. Fang, C.~Wang, M.~Gou, and C.~Lu, ``Graspnet-1billion: A large-scale benchmark for general object grasping,'' in \emph{Proceedings of the IEEE/CVF conference on computer vision and pattern recognition}, 2020, pp. 11\,444--11\,453.

\bibitem{song2020improved}
Y.~Song and S.~Ermon, ``Improved techniques for training score-based generative models,'' \emph{Advances in neural information processing systems}, vol.~33, pp. 12\,438--12\,448, 2020.

\bibitem{vaswani2017attention}
A.~Vaswani, ``Attention is all you need,'' \emph{Advances in Neural Information Processing Systems}, 2017.

\bibitem{rombach2022high}
R.~Rombach, A.~Blattmann, D.~Lorenz, P.~Esser, and B.~Ommer, ``High-resolution image synthesis with latent diffusion models,'' in \emph{Proceedings of the IEEE/CVF conference on computer vision and pattern recognition}, 2022, pp. 10\,684--10\,695.

\bibitem{xu2021seeing}
H.~Xu, Y.~R. Wang, S.~Eppel, A.~Aspuru-Guzik, F.~Shkurti, and A.~Garg, ``Seeing glass: joint point cloud and depth completion for transparent objects,'' \emph{arXiv preprint arXiv:2110.00087}, 2021.

\bibitem{zhu2021rgb}
L.~Zhu, A.~Mousavian, Y.~Xiang, H.~Mazhar, J.~van Eenbergen, S.~Debnath, and D.~Fox, ``Rgb-d local implicit function for depth completion of transparent objects,'' in \emph{Proceedings of the IEEE/CVF Conference on Computer Vision and Pattern Recognition}, 2021, pp. 4649--4658.

\bibitem{coumans2016pybullet}
E.~Coumans and Y.~Bai, ``Pybullet, a python module for physics simulation for games, robotics and machine learning,'' 2016.

\end{thebibliography}

\end{document}